\documentclass[12pt]{ouparticle}
\usepackage{hyperref}
\usepackage{cleveref}
\begin{document}

\title{Adma: A Flexible Loss Function for Neural Networks}

\author{%
\name{Aditya Shrivastava\thanks{Author's note: I am currently a final year undergraduate at Nirma University. I've tried to consolidate my year's worth of research here. Due to lack of availability of time, I couldn't get it published. The paper has room for more substantiation and might also be a bit difficult to follow. But, I've tried my best to justify the imperative and plausible future directions, especially in the conclusion section. Please feel free to reach me out regarding any concerns or improvements. I am also personally available at: \href{mailto:adityashrivastava2799@gmail.com}{adityashrivastava2799@gmail.com}.}}
\address{Institute of Technology, Nirma University}
\email{\href{mailto:17bit014@nirmauni.ac.in}{17bit014@nirmauni.ac.in}}}

\abstract{Highly increased interest in Artificial Neural Networks (ANNs) have resulted in impressively wide-ranging improvements in its structure. In this work, we come up with the idea that instead of static plugins that the currently available loss functions are, they should by default be flexible in nature.  A flexible loss function can be a more insightful navigator for neural networks leading to higher convergence rates and therefore reaching the optimum accuracy more quickly. The insights to help decide the degree of flexibility can be derived from the complexity of ANNs, the data distribution, selection of hyper-parameters and so on. In the wake of this, we introduce a novel flexible loss function for neural networks. The function is shown to characterize a range of fundamentally unique properties from which, much of the properties of other loss functions are only a subset and varying the flexibility parameter in the function allows it to emulate the loss curves and the learning behavior of prevalent static loss functions. The extensive experimentation performed with the loss function demonstrates that it is able to give state-of-the-art performance on selected data sets. Thus, in all the idea of flexibility itself and the proposed function built upon it carry the potential to open to a new interesting chapter in deep learning research.



}

\date{\today}

\keywords{Loss Functions; Flexible learning method; Neural Networks; Classification and Deep Learning.}

\maketitle

\section{Introduction}
{A}{}rtificial Neural Networks (ANNs) have undergone a revolution from their initial stages of research \cite{C57, C61, C58, C59, C62, C60} compared to the highly extensible and customisable structure that they have now\cite{C63, C64, C65, C66, C11}. Deep Learning community has been successful in extending its framework to solve a wide variety of problems, as they have found their applications ranging from visual classification tasks \cite{C1, C2, C3, C4} to natural language processing \cite{C5, C6, C7, C8, C9, C10} and various other miscellaneous tasks \cite{C67, C11, C12, C13}. The field has almost faced a resurrection since \cite{C1} and from then on, the researchers have continued to produce meaningful modifications on each part of the ANN architecture to improve upon the state-of-the-art methods.
Such great flexibility in designing each element of the entire structure separately is one of the main reasons of Deep Learning being so successful. With that being said, it is also important to mention about the years of research that has been done to improve upon each and every component of the neural networks, be it significant or non-significant. For eg. imposing changes at the data preprocessing stage i.e. prior to inputting the data in the neural network such as normalization \cite{C14}, data augmentation \cite{C1, C15} or using large-scale training data \cite{C16}. Further, the important techniques to improve the performance that engineer the internal structure of the neural networks include incorporating deeper structures \cite{C16, C15}, changing stride sizes \cite{C17}, utilizing different pooling types \cite{C18}, regularization techniques \cite{C20, C21, C22}, activation functions \cite{C19, C23, C24, C25} and in optimizing the networks \cite{C26, C27}. However, even though after such breakthroughs following the desire to achieve modular performance, one element is still kept unviolated--loss function. For almost 30 years, since the works \cite{C68, C69, C70} have proposed to use cross-entropy as a loss function, it has universally remained the default choice of loss function for most classification and prediction tasks involving deep learning. And after that, neither there is much work done on the improvement of prevalent loss functions as compared to other parts of the ANN architecture nor is much variety being developed in the choice of loss functions.


The loss functions form one of the most important characteristic in the design of ANNs \cite{C28} whatsoever. Moreover, the luxation of the component of loss function imposes nominal intervention in the designed model as compared to the other methods that often demand exhaustive changeover in the entire design of neural networks. Still, the research work on improvement of loss functions is relatively dwarf as compared to the other components of the network. During the earlier stages of the research, Mean Squared Error (MSE) loss had been the choice of loss function \cite{C29, C30, C31}. It was originally used for the regression problems before being derived for neural networks using maximum likelihood principle by assuming the target data to be normalized \cite{C32}. The problem observed with this error function is in the assumption of normalized nature of the data. The real world data is not always observed to be normalized. Also, it is often desirable while designing a neural network that its cost function gradients are of the considerable magnitude so that it can learn properly. Unfortunately, the networks start producing very small gradients when MSE is used as a cost function. And this often causes the learning to saturate quickly. It is also sometimes observed that MSE function fails a model from training even when it produces highly incorrect outputs \cite{C28}. This had lead to move to cross-entropy as a choice of loss function. All of the properties for MSE can also be applied to this function as well like this function can also be derived using the principle of maximum likelihood. There are several other reasons as well that motivate to use cross entropy function. One of the well established reason is that when it is coupled with a commonly used softmax output layer, it undoes those exponentialized output units. Further, the properties of cross-entropy function can be even thought to be closely related to the properties of Kullback-Leibner divergence that describes the differences between two probability distributions. Hence, minimizing the cross entropy function would result in minimizing the KL divergence. Due to these and many other reasons, in the recent years, cross-entropy has been the persistent choice of loss function. Some of the literature have mentioned using other functions such as Hinge Loss when it comes to performing the classification task \cite{C33, C34, C35, C36}. In addition to these popularly used loss function there are also few other complex loss functions proposed such as triplet loss \cite{C37}, contrastive loss \cite{C38} etc. Triplet loss requires 3 input samples to be injected simultaneously whereas the contrastive loss inputs two samples simultaneously and requires the features of same class to be as similar as possible. These loss functions, even after having such pervasive intuition behind their usage, seem to be missing out on one of the issues i.e. these functions do not tend to be implicitly flexible about the amount of information that has to be back-propagated depending upon both the data on which the model is trained and the design of the network model. To this end, we propose a new adaptable loss function that is built upon the core idea of flexibility. Subsequently, upon analysis and experimental validation, we justify that the function not only comprises of the fundamental properties that are seen in other cost functions but also presents with some unique and advantageous properties that give us interesting directions to work upon in future.

The key intuition behind this function is that the Neural Network has to be flexible in learning and therefore in estimating its error as well, depending upon the characteristics proclaimed by the data and learning environment. Such adaptability to the characteristics of the data can have several upsides. Because the loss that is calculated can be scalable, its difficulty can be adjusted that can force the network to learn only the most noteworthy and discriminative features and the features contributing the most to the corresponding class. This can lead the neural network to be more immune to overfitting problem, show higher convergence rate and therefore may also require lesser data to train. 

`One may surely find upon investigation that the field of Deep Learning has every now and then, drawn inspiration from the learnings of Neuroscience \cite{C40}. Therefore, our intuition is also backed by the learnings of neuro-psychology, a branch of neuroscience. We understand that our brain has different approximation for our mistakes under different events. One trivial instance for this can be the difference in manually diagnosing and predicting images from MNIST and CIFAR dataset. On interpreting an image containing a handwritten 6 as being an 8, the mistake can be easily excused or pardoned to clumsy handwriting. On the other hand, the error in incorrectly classifying a bird as an airplane is taken much more seriously, almost to an extent that such mistake never happens again. It can be argued in this case that difference between learning in both the cases shall be implicit adhering to difference in their respective pixel arrangements of MNIST and CIFAR datasets. However, we suggest that in manual case the difference in errors and resultant learning from it is more because of difference in the respective categories of the datasets. Similarly, we sometimes fail to recognize the same image correctly when under different mindsets or moods. This suggest that our decisions and perceptions are comprehensive of various factors and therefore are flexible and adaptable in different conditions. This inspires to make the loss functions responsible for calculating the errors and making decisions in a deep learning model to also become flexible and adaptable in their tasks as well.



As a result, the amount of information that is flown backward in the form of a correction of a mistake becomes flexible as well. Therefore, in this work we propose a novel flexible loss function for neural networks. Using this function, we show how the flexibility of the loss curve of the function can be adjusted to improve the performance as such--reducing the fluctuation in learning, attaining higher convergence rates and so on. This can altogether help in achieving the state-of-the-art performance in a more plausible manner. 



Commonly, whenever starting out with the novel, nature inspired idea, it becomes a need to mention the current state-of-the-art methodologies and justify how the proposed idea, even after differing from them, shall fit well into scenario. Then, the intuition behind the idea needs to be described and finally, the idea shall be proved through experimental validation \cite{C39}. Thus, we also follow the same chronology in this paper. 

The paper is further organized as follows: section \ref{second} reviews key concepts behind the loss functions and work done to add adaptability in the loss functions. Then, in Section \ref{third}, we propose the loss function and describe its various properties. In Section \ref{four} the technicalities of experimental validation is specified and consecutively in Section \ref{fifth} a discussion is presented about the results achieved through experimental validation performed. Lastly, in Section \ref{six}, we make the concluding remarks.



\section{Background} \label{second}
\subsection{Preliminaries \& Notation}
In general, a problem of k-class classification is considered. Here the input feature space $ \mathbb{X} \subset R^d$ is classified as belonging to one of the label space $ \mathbb{Y}=\{ 1, \dots, c \}$ and the classifier function $f(x;\theta)$ is used to map the input feature space to the classes $f : \mathbb{X} \rightarrow \mathbb{Y}$. The function $f$ predicts the posterior probabilities for each class i.e. each element of set $\mathbb{Y}$. The goal, according to the Maximum Likelihood (ML) principle is to estimate the optimal set of parameters $\theta$ for which the expectation for the classifer $f(X;\theta)$ of generating the true set of probability distribution $\mathbb{Y}$ is maximum which means, 
\begin{equation}\theta^*=\underset{\theta}{arg\,max}\,\mathbb{E}_{x\sim y}f(X;\theta)\end{equation}
It can be interpreted that maximizing the expectation of the classifier function $f(X;\theta)$'s prediction to be equivalent to the true label $y$ is to minimize the the dissimilarity between the two distribution. This quantification of the difference between the model output $f(x;\theta)$ and true label is done using the loss function.\begin{equation}\mathcal{L}(f(X;\theta), y)\end{equation}
\begin{equation}\sum_{j=1}^{c}\mathcal{L}(f(x;\theta), j)\end{equation}
The end goal, however is to find an optimal function $f^*$ such that the dissimilarity is at its minimum. 
\begin{equation}
f^*=\underset{f}{arg\,min}\,\mathbb{E}_{x\sim y}\mathcal{L}(f(X;\theta), y) \label{eq4}   
\end{equation}There different loss functions proposed and each of them have their own properties and they quantify the difference in their own way. The good design of the loss function is thus an important issue to ponder upon for appropriate learning by the neural network.

\subsection{Related Work}



The core agenda of the proposed work is to introduce adaptability or flexibility during the quantification of loss. Hence, in this subsection we review the efforts that have been made to extend the calculation of loss done by the cost functions to be more adaptable. The most common method is to couple the loss given by the model with some a regularization method such as parameter norm penalty that can act as constraint. Such constraint can penalize the capacity of the model causing it to learn the most important features first \cite{C28}. Further, when it comes to adding a constraint on the model, providing a neural network with the symbolic knowledge has also proved to be a satisfactory choice \cite{C41}. Such symbolic knowledge is interpreted as a constraint by the model which can guide the model even more fittingly especially under weak supervision. 

Another way to enable the model learning to be more adaptable is to appropriately characterize the separability between the learned features of the input data \cite{C42}. In this method, a combination of the last fully connected layer, softmax function and cross-entropy is done and the functionality of angular similarity between the input sample and parameters of the corresponding class is added to explicitly encourage the intra-class compactness and inter-class seperability of the learned features. The combination is termed as softmax loss and is justified to be a large-margin softmax (L-Softmax) loss by adding the flexibility of inter-class angular margin constraint.

Further, the performance of the model is also often suffered due to the noise present in the data in form of mislabelled examples. Recent efforts are done to make the loss function generalized from the noise-robustness viewpoint \cite{C44, C43, C45}. Noise robustness of the Mean Absolute Error (MAE) is proved theoretically by Ghosh et al.\cite{C43}. They have mentioned that since there is no presence of the term $1/f(x;\theta)$ in its gradient, it would treat every sample equally which makes it more robust to noisy label. Nevertheless, it can be demonstrated empirically that this can lead to longer training procedure. Also, the stochasticity that results without implicit weighting can cause the learning to be more difficult. On the other hand, the cross-entropy has implicit weighting functionality but can not be replaced by MAE when it comes to making the learning noise-robust because of its nonsymmetric and unbounded nature. Therefore, Zhang et al. \cite{C45} proposes to exploit both implicit weighting feature of cross-entropy and noise-robustness of MAE using negative of Box-Cox transformation as loss function. By adjusting a single parameter, it can interchangeably act as both MAE and cross-entropy. 

Since, the cross-entropy itself is sensitive to noise, recently Amid et al. \cite{C46} introduced a parameter called "temperature" in both the final softmax output layer and in the cross-entropy loss. By tuning both "temperature" parameters, they make the training all the more robust to noise. When it comes to using various loss function interchangeably, an unconventional approach is made by Wu et al. \cite{C47} where a meta learning concept is developed and explained as teacher-student relation. The parametric model (teacher) is trained to output the loss function that is most appropriate for the machine learning model (student).

One work that deserves to be particularly addressed is that of Focal Loss \cite{C71}. The function proposes flexibility as means of focusing more on harder misclassified samples. We spot some difference between the objectives of the function proposed here and focal loss. First, even though the function's curves are shown to be flexible, its limits remain unaltered. This means there is still one side of function asymptoting to infinity. Hence, there is still a possibility of loss leakage phenomenon leading to hit NaN loss while learning. Second, since the function is based on cross-entropy its acceptance is also partially backed by the intuition that log function undoes the exponential units of softmax to allow back-propagating its gradients linearly. We show with our loss function that even further exponentializing the already exponentialized outputs of softmax can achieve state-of-the-art performance. Third, it importance is less justified for straightforward image classification tasks and lastly this function as well is based on extending the design of cross-entropy to incorporate flexibility in it. 

Hence, all of the above described approaches have one thing in common i.e. all of them necessitate performance improvement upon the respective loss function itself or on the combination with some other element of the model. In a nutshell, it can be inferred that none of the functions allow the model to learn flexibly and robustly when in its primary structure. Thus, unlike any of the previous approaches described, we introduce a loss function that has flexibility as its core idea. And as we shall see, this flexibility property of the loss function shall allow the function to also emulate the behaviour of other loss functions.

\section{Adma: An Adaptable Loss Function} \label{third}
Following the notation described in the Background, the exact definition of the Adma function is:
\begin{equation}
    \mathcal{L}(f(x;\theta), y)= y(e-e^{f(x;\theta)^a})\label{eq5}
\end{equation}
where the literal $a$ is the \textit{scaling factor} that controls the flexibility while quantifying loss. Consequently, substituting the value of eq. \ref{eq5} in eq. \ref{eq4}, the final optimized function $f^*$ or set of parameters $\theta^*$ can be obtained as,
\begin{equation}
f^*=\underset{f}{arg\,min}\,y(e-e^{f(x;\theta)^a})
\end{equation}
\begin{equation}
\theta^*=\underset{\theta}{arg\,min}\,y(e-e^{f(x;\theta)^a})
\end{equation}

From our experiments, it is observed that the model yields best performance for when $a \in [0.0, 0.5]$. Also, the function satisfies the following basic mathematical requirements with $a$ in this range which otherwise seem to be violated:
\begin{itemize}
\item \textbf{Well-defined} range from $[0, e-1]$.
\item \textbf{Non-negative} in this range.
\item \textbf{Differentiable} domain.
\item \textbf{Monotonically} decreasing nature.
\end{itemize}

Therefore, we make implicit consideration to analyse the properties of the function supposing $a$ in this range. Further, the loss function also assumes the function $f(x;\theta)$ to be the set of posterior probabilities for each of the classes produced by the model. 

The function, clearly has a well-defined range. This can be attributed to being more shielded to numerical instability such as hitting \textit{NaN} loss, starting to diverge quickly etc. Since, for the same value of output, on increasing the value of scaling factor, the effective exponential output diminishes, the scaling factor can be indirectly thought as a regularization parameter to limit the capacity of the model. This may force the to learn the most reasonable features first and prevent it from becoming overfitted.

\begin{figure}[H]
\centering
\includegraphics[scale=0.28]{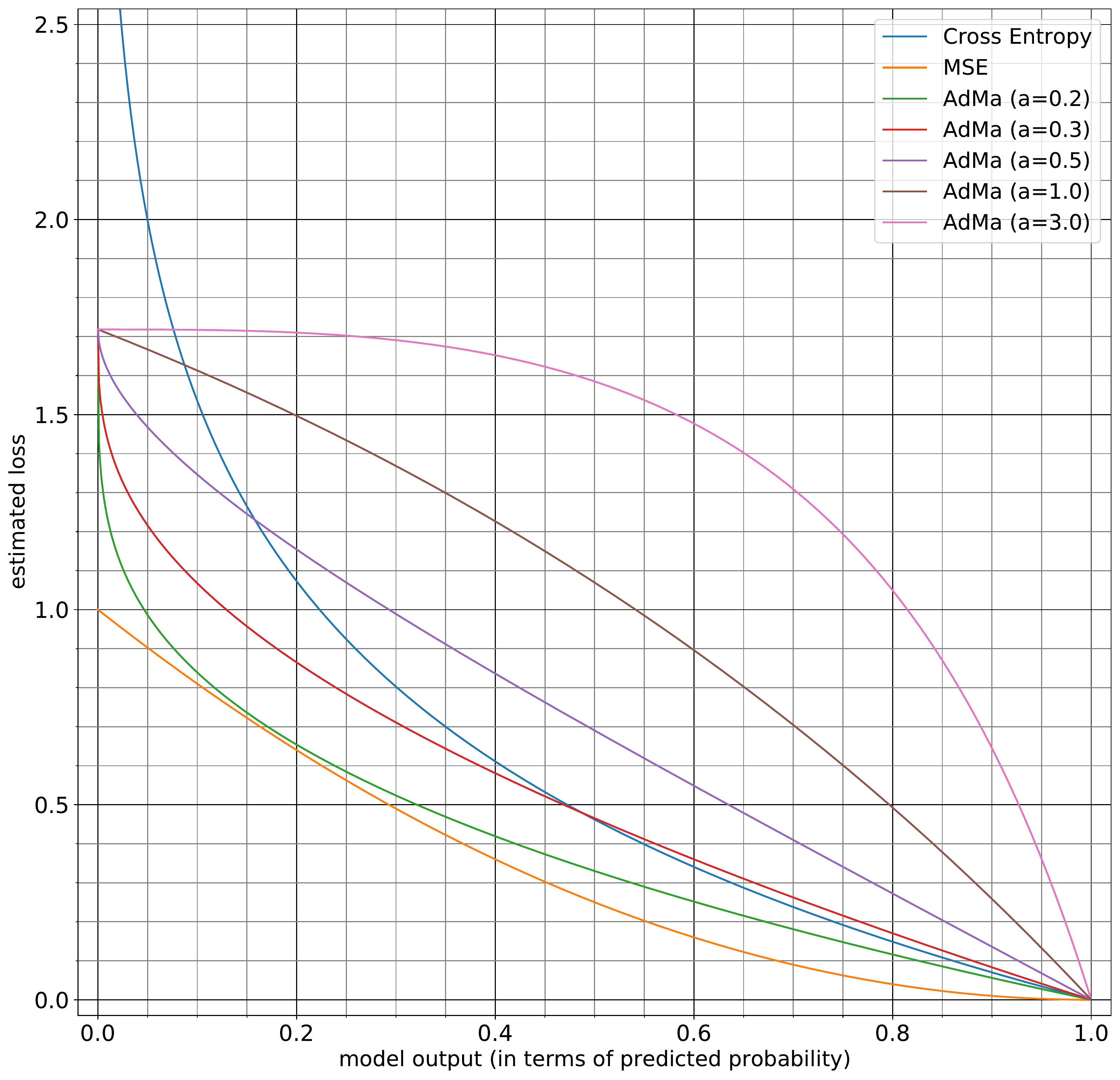}
\caption{The loss curve of various loss functions. Here, the flexible nature of Adma clearly shows to match the loss curve of Cross-Entropy}
\label{fig1}
\end{figure}

The function's flexibility in the shape and slope is clearly demonstrated in the Figure \ref{fig1}. Here, it is shown that for the same range of values the Adma loss function can show a high flexibility in quantification of loss for the model's output.

It can be deduced from the figure that the function can even imitate the shape equivalent to that of cross-entropy for some value of $a$. However, as we see that on increasing the value of $a$ beyond 0.5, the function begins to lose its convexity and starts showing a concave nature.

\subsection{Gradient Analysis}

The gradient of the \eqref{eq5} is,

\begin{equation}
\frac{\partial\mathcal{L}(f(x;\theta), y)}{\partial\theta}= -a{f(x;\theta)^{a-1}}(e^{f(x;\theta)^a})\cdot\nabla_{\theta}f(x;\theta)
\end{equation}
The explicit components of scaling factor, predicted value and the exponential function are observed in the gradient from which some of the evident advantages can be directly inferred.
\begin{enumerate}
\item The property of contrast for the scaling parameter $a$ as observed in the function's expression is also observed here amongst the co-efficients of the gradient, 

\begin{equation}
a \propto \frac{1}{f(x;\theta)^{a-1}(e^{f(x;\theta)^a})}
\end{equation}
This suggests that there has to be an optimal value of $a$ for training. Both an excessively increased or decreased value of $a$ can cause cumbersome learning in the model.

\item Eq. \ref{eq5} can be rewritten as,

\begin{equation}
\frac{\partial\mathcal{L}(f(x;\theta), y)}{\partial\theta}= -a{\frac{1}{f(x;\theta)^{1-a}}}e^{f(x;\theta)^a}\cdot\nabla_{\theta}f(x;\theta)
\label{eq10}
\end{equation}

\begin{equation}
\implies
\frac{\partial\mathcal{L}(f(x;\theta), y)}{\partial\theta} \propto \frac{1}{f(x;\theta)}
\label{eq11}
\end{equation}

For the presumed range of values of $a$, the co-efficient of the model's output (present in R.H.S of eq. \ref{eq10}), is inversely proportional to the gradient of the function (refer eq. \ref{eq11}). Therefore, this function also has the implicit weighting scheme analogous to the one present in the cross-entropy function, which means that the term $1/f(x;\theta)^{1-a}$ would comparatively weigh more on the samples that match weakly with the true label and conversely weigh less on the samples matching strongly with the ground truth.

\item The gradient has the component $e^{f(x;\theta)^a}$ where  
\begin{equation}
f(x;\theta)^a \in [0, 1] \implies e^{f(x;\theta)^a} \in [1, e]
\end{equation}
So, the gradients during backpropagation would always be amplified and in turn satisfy one of the most important criteria for efficient design of the neural network i.e. having adequately large gradients so that the function can serve as coherent guide for learning without saturating quickly.

\end{enumerate}

Hence, the above theoretical discussion advocates that the proposed function besides being flexible can also emulate the behavior observed in commonly used loss functions to a notable extent. Therefore, we now move forward to evaluate the model's performance by exploiting the proposed function on various datasets and diverse settings.

\section{Experimental Setup} \label{four}


We provide an empirical study to experimentally validate the capabilities of the proposed loss function. In this section, we describe all of the configurations and conditions under which our loss function was tested. To ensure a genuine comparision,  we perform all tests by replacing only the loss function and keeping rest of the configurations identical.

\subsection{Architecture Design}
\begin{figure}[!htbp]
\centering
\includegraphics[width=0.5\columnwidth]{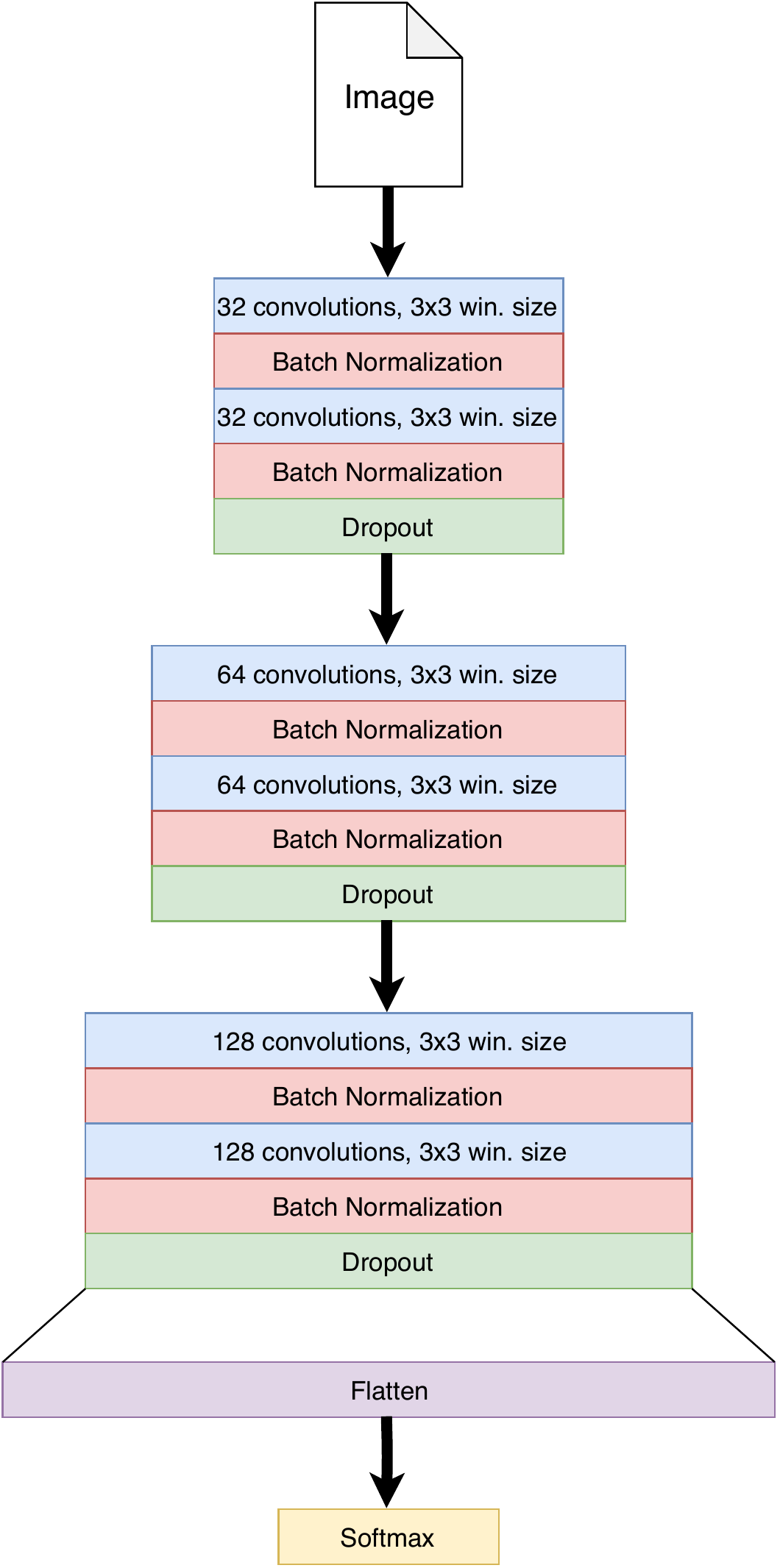}
\caption{Layout of a custom built CNN (ConvNet)}
\label{fig2}
\end{figure}

To perform the tests with our loss function, we consider a manually designed convolutional neural network model. The reason being the need to explore the settings or hyper-parameters under which the function tends to perform best. As described in Figure \ref{fig2}, the CNN architecture basically has 3 blocks, with each block having exponentially higher number of convolutions than the previous one. The model has an 'ELU' activation unit. Further, the structure of this baseline architecture is negotiated by editing its components like activation function, adjusting dropout rate, introducing regularization etc. The performance change is evaluated corresponding to such changes under each of the considered loss functions. In the rest of the paper, this architecture is referred simply as ConvNet.

Lastly, just to confirm upon the results, we perform a revised round of experiments with a pretrained model--ResNet34. Thus, the model shall only be used for benchmarking purpose and therefore shall not have to go through the extensive experimentation similar to that of manually designed ConvNet.

\subsection{Dataset Selection}
We evaluate our models performance on a total of 8 datasets: MNIST\cite{C49}, Fashion-MNIST \cite{C50}, CIFAR10 \cite{C51}, CIFAR100 \cite{C51}, Flower Recognition \cite{C48}, Skin Cancer \cite{C52}, Street View House Number Dataset \cite{C53} and Dogs vs. Cats \cite{C54}.
Following passage gives description about the characteristics of the chosen datasets and the baseline settings under which they are trained.  
\subsubsection{MNIST} 
We start by considering the most commonly used image classification dataset. It consists of a total of 70,000 28x28 greyscale images of handwritten digits belonging to a total of 10 classes. From the total dataset 50,000 images for training, 10,000 for validation and last 10,000 for testing. Since this is the most standard dataset having elementary characteristics, models usually end up achieving almost a 100\% accuracy. Therefore, we only run our experiments for 10 iterations where each iteration lasts for 50 epochs. The dataset, however, is augmented by allowing rotation range of $30^{\circ}$ and shifting range of 20\%. 

\subsubsection{FASHION-MNIST}
This dataset consists of Zalando's downsampled low-resolution article images. This dataset aims to serve as a more challenging yet as a standard replacement for MNIST.
Other than the classes of this dataset, all of the other characteristics are similar to that of MNIST dataset. Therefore, we apply identical pre-processing and training setup as in MNIST except that the model is trained for 75 epochs.

\subsubsection{CIFAR-10}
The CIFAR-10 dataset is formed as a small subset of the original 80 million tiny images dataset containing labels.
It is comparatively more sophisticated than the previous two datasets. Here we employ a standard subset of the entire dataset for our experiments. It consists of 32x32 colored RGB images belonging to 10 classes and split into 50,000 train and 10,000 validation images. Our experiments on this dataset run for 20 iterations where each iteration in-turn runs for 125 epochs.
Data normalization, horizontal random flip, warping, height and width shifting upto 20\% were performed as part of data preprocessing and augmentation. The batch-size is of 64.

\subsubsection{CIFAR-100}
We also evaluate our loss function on the original CIFAR-100 dataset. It consists a total of 50,000 32x32 training images over 100 labelled categories and another 10,000 images for validation. We train the our models on this dataset for 20 iterations and 170 epochs. Since the dataset itself has quite a few classes, we impose a relatively low data augmentation compared to CIFAR10 dataset. The augmentation includes $15^{\circ}$ rotation range, 10\% of horizontal as well as vertical shift and horozontal random flip.

\subsubsection{Flower Recognition}
This dataset consists of colored images of 5 different flowers namely, chamomile, tulip, rose, sunflower, dandelion. It has a total of 4,242 flower images with approximately 800 images of each of the flowers. The dataset originally has a resolution of 320x240. However, since it is quite difficult to train the model on such high resolution images with the available computational resources, they have been sized to 150x150 resolution. The train-test split was kept at 3:1 ratio i.e. from each class around 600 samples were utilized for training and the rest 200 were left for validation. Further, the ConvNet model is employed for training with a batch-size of 64 for 20 iterations and each iteration lasting 150 epochs. The learning rate was $4\times10^{-4}$.

\subsubsection{Skin Cancer}
This dataset has 10,015 dermatoscopic images. The images belong to either of the seven types of cancer: Actinic keratoses (akiec), basal cell carcinoma (bcc), benign keratosis-like lesions (bkl), dermatofibroma (df), melanoma (mel), melanocytic nevi (nv) and vascular lesions (vasc). The dataset originally had the images of the shape 450x600x3. Due to computational constraints, we resize the images to shape of 75x100. Apart from this, the rest of the training procedure is identical to the Flower Recognition dataset.

\subsubsection{Street View House Numbers}
The Street View House Numbers (SVHN) dataset consists of RGB images of housenumbers collected by google street view. The images consist of multiple numbers and goal is to correctly identify the digit at the center of the image. The dataset has a total of 73,257 images in the training set and 26,032 images in test set. Other than these, it also has additional 5,31,131 less difficult examples. Since, the task is to identify only a single number, we ignore the additional examples and only consider the standard training and testing set for our examples. 

\subsubsection{Dogs vs. Cats}
Lastly, as a binary classification problem we test the loss function on the classical dogs vs cats dataset. The dataset originally consists of over 3 million images. However, for validation, only a subset of 25,000 images is considered, having 12,500 images belonging to each of the classes. All of the images are cropped to a 150x150 resolution. Since this is only a binary classification problem, the model is run for only 50 epochs. 

\subsection{Other Settings}
Here we describe all of the details other than the models and datasets that were required in our training setup. For the optimization of our models, we consider SGD and Adam as they are the most widely used optimizers. And also, to some extent they possess the beneficial properties of the other optimizers. However, both the optimizers have there own pros and cons and there aren't any grounded proofs mentioning one optimizer being better than another under all circumstance. The experimental difference between the two is that Adam being adaptive, has higher convergence rate and requires less experimentation, whereas, SGD tends to have a better generalization capability for the models that are over-parametrized \cite{C55}. Therefore, we perform the experiments sequentially considering both these optimizers. Further, the results are validated in both the presence and absence of regularization. Generally, a high variance is introduced in the model due to such regularization. Therefore, just to validate the robustness of our loss function as compared to others, we evaluate the performance of the model under L2 regularization with a weight decay of $10^{-4}$ employed in each of the convolutional layers. Its impact is discussed in the discussed in Section \ref{fifth}. Lastly, the results of the function are compared with the most commonly used functions i.e. Categorical Cross-Entropy (CCE), Mean Squared Error (MSE) and Squared Hinge. As an evaluation parameter, we report the average validation accuracies (VA) obtained at the end of every iteration in case of each of the datasets corresponding to each of the models.

\section{Results \& Discussion} \label{fifth}
\begin{figure*}[!ht]
\begin{multicols}{4}
    
    \includegraphics[width=\linewidth]{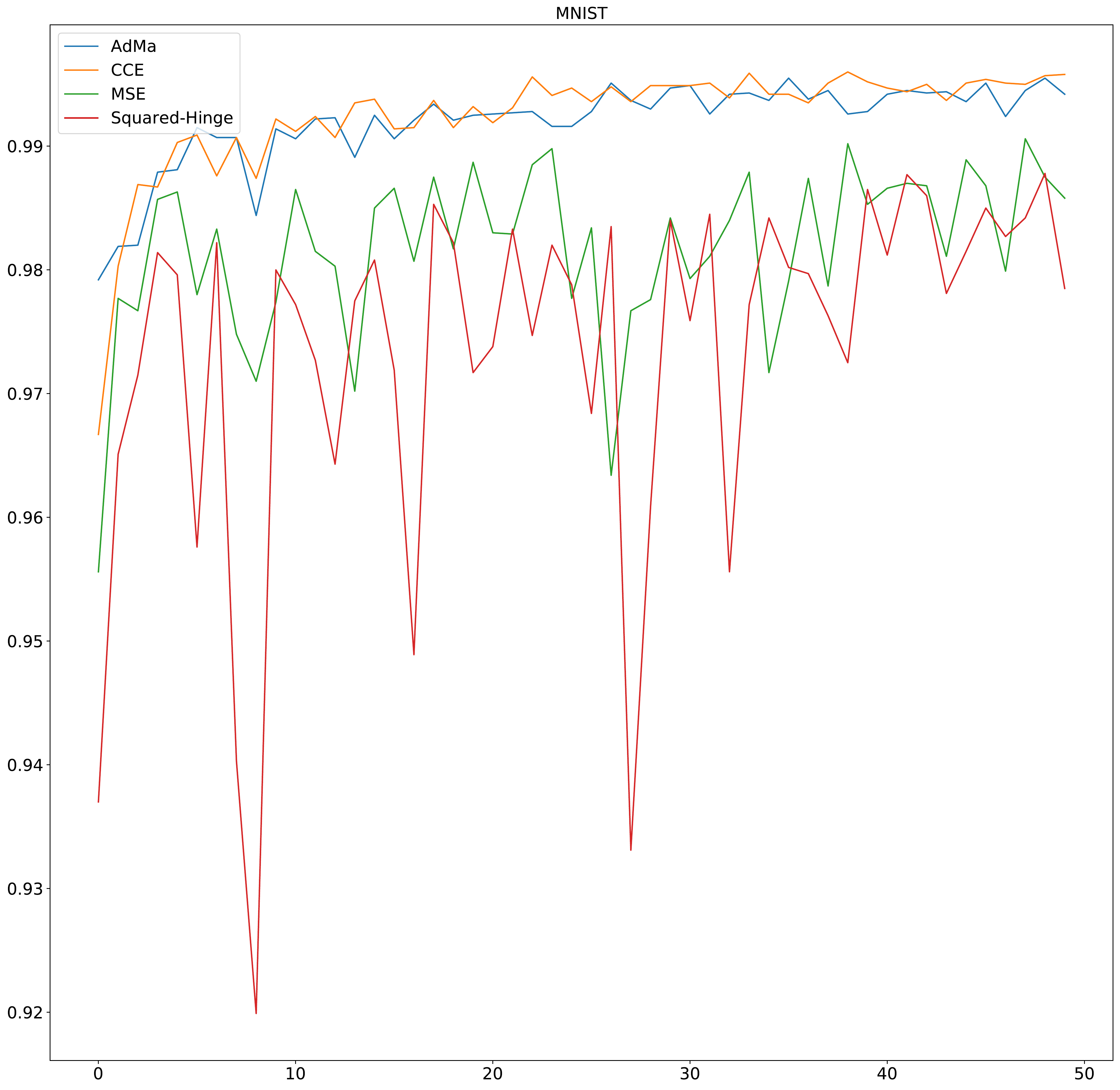}\par
    \includegraphics[width=\linewidth]{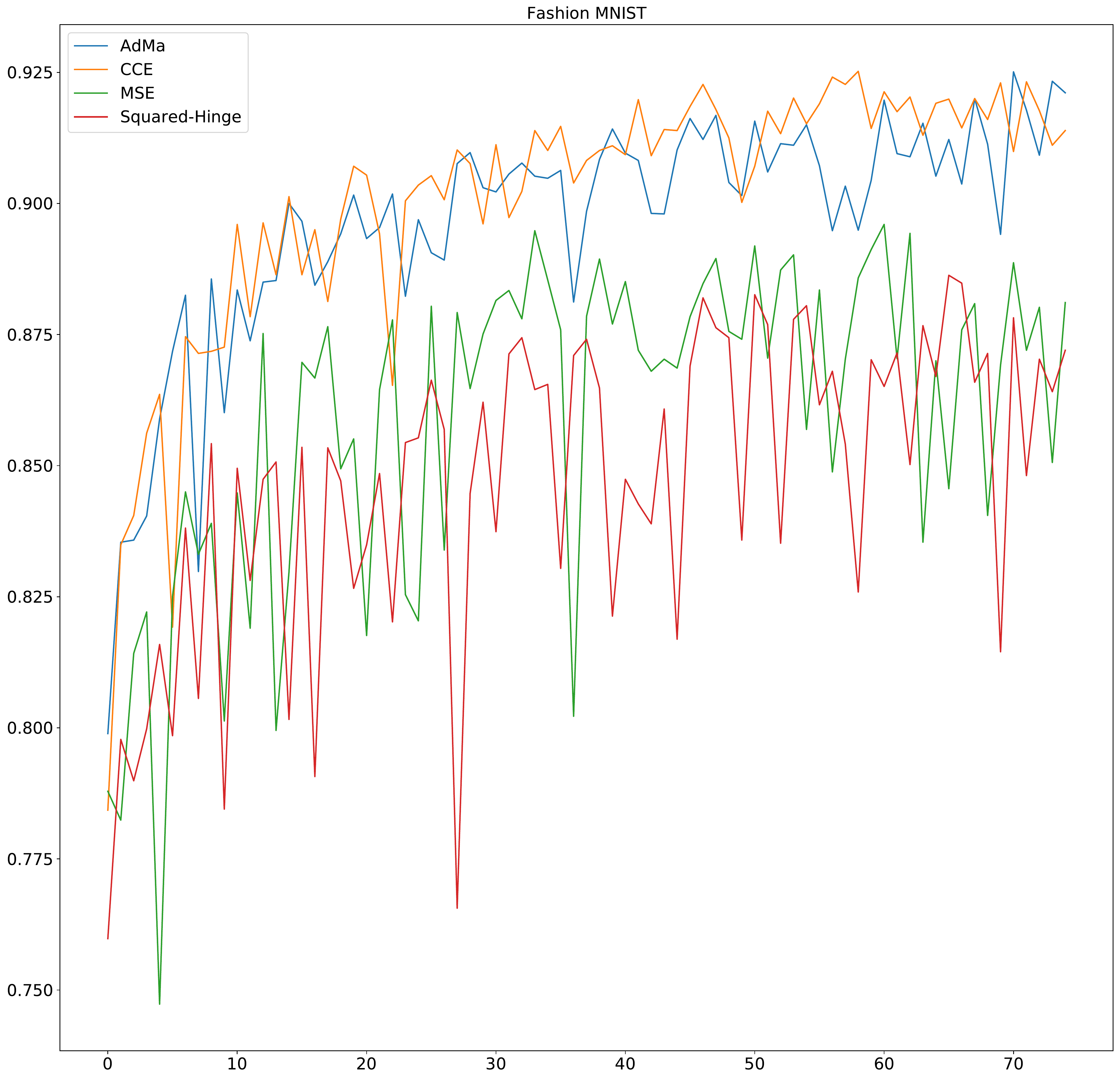}\par
    \includegraphics[width=\linewidth]{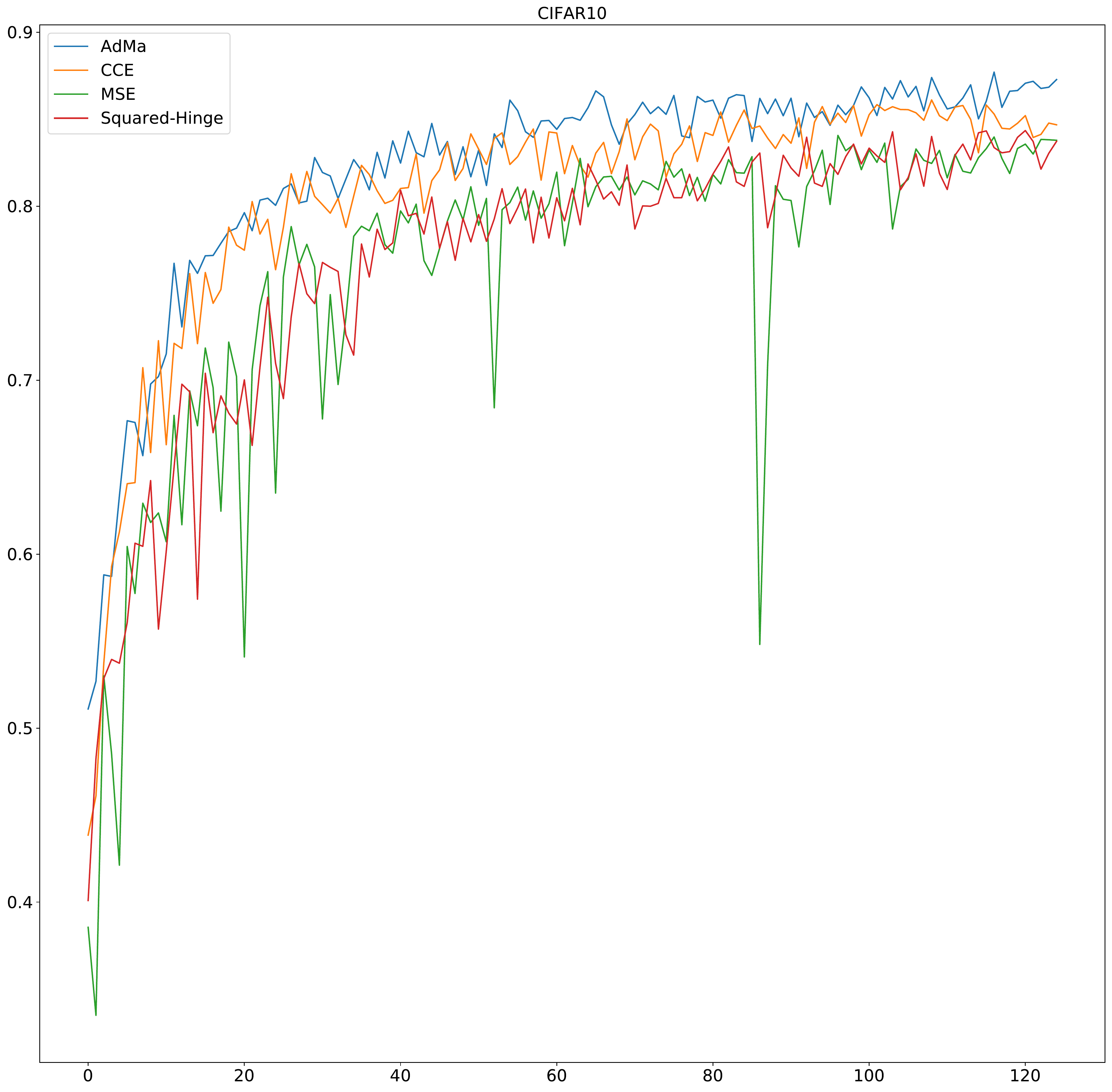}\par 
    \includegraphics[width=\linewidth]{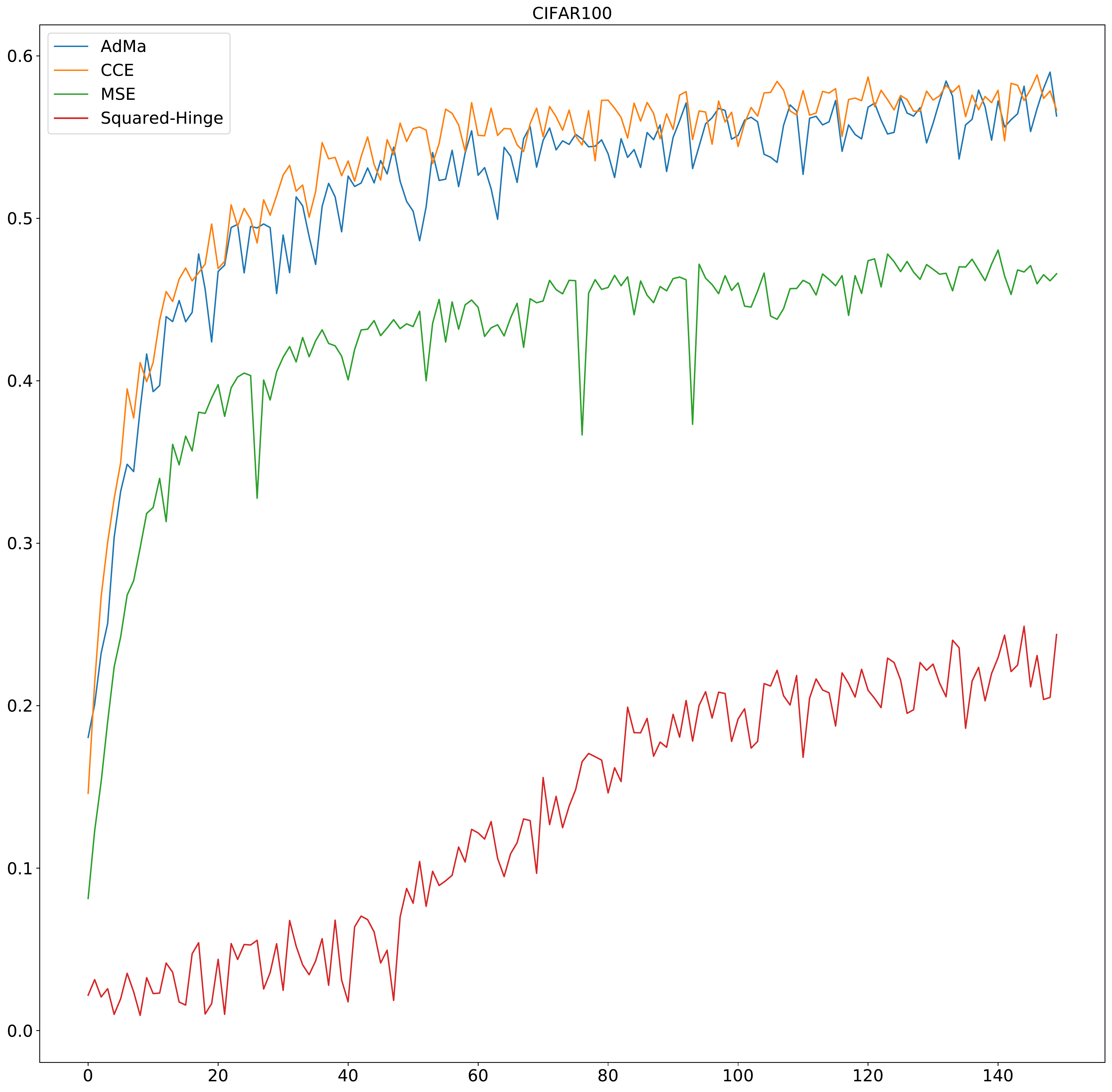}\par 
\end{multicols}
\begin{multicols}{4}
    
    \includegraphics[width=\linewidth]{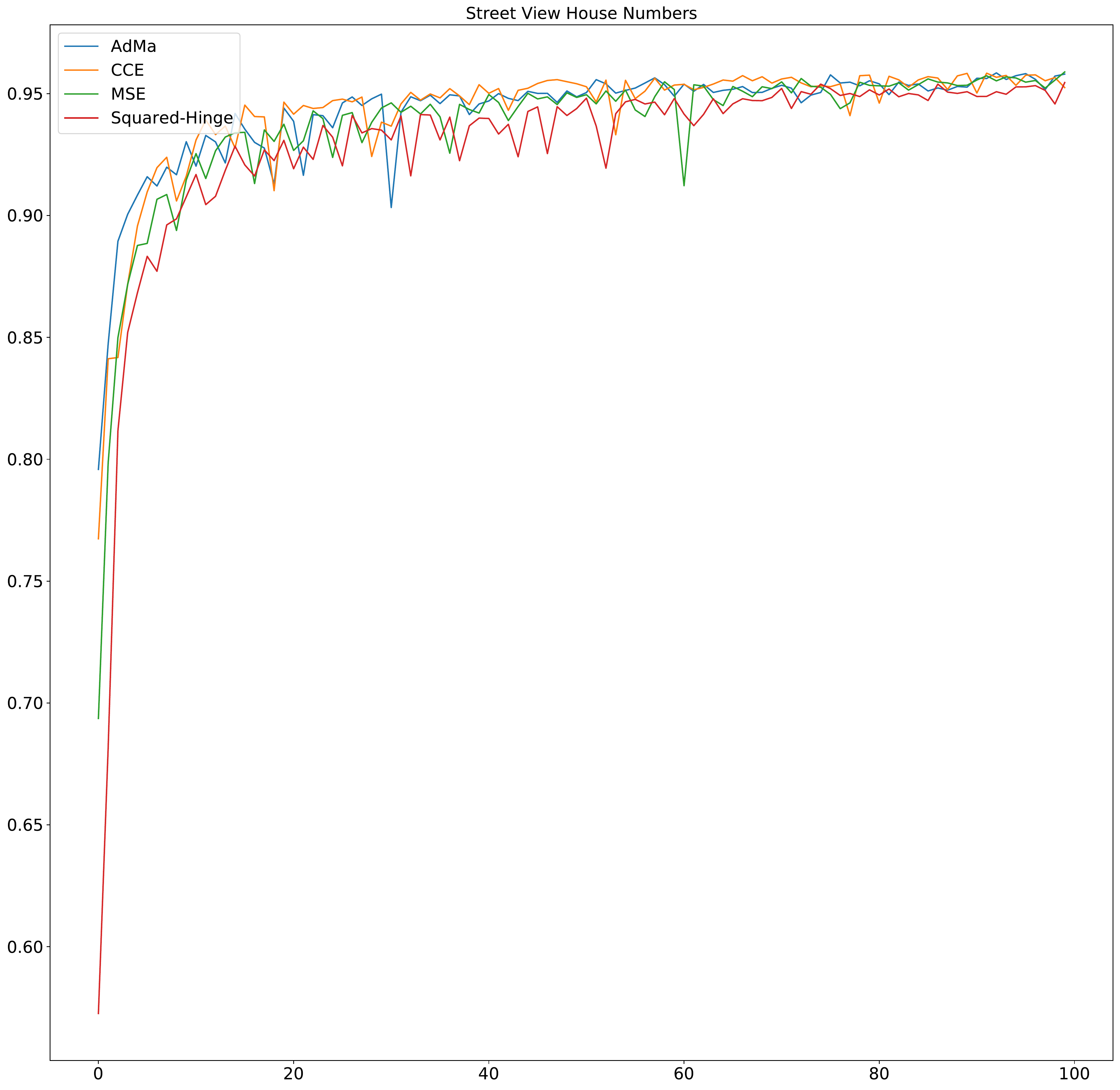}\par
    \includegraphics[width=\linewidth]{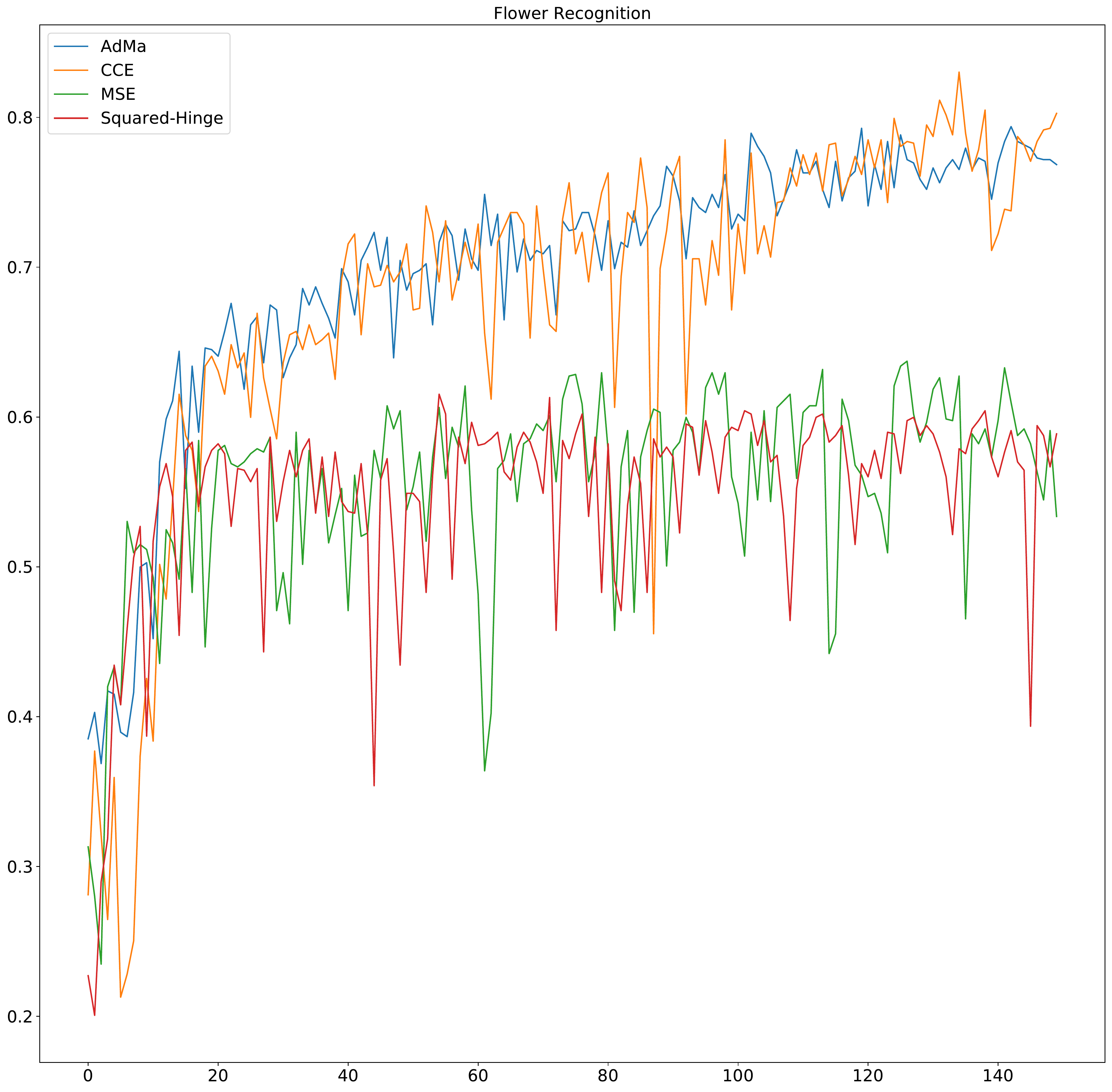}\par
    \includegraphics[width=\linewidth]{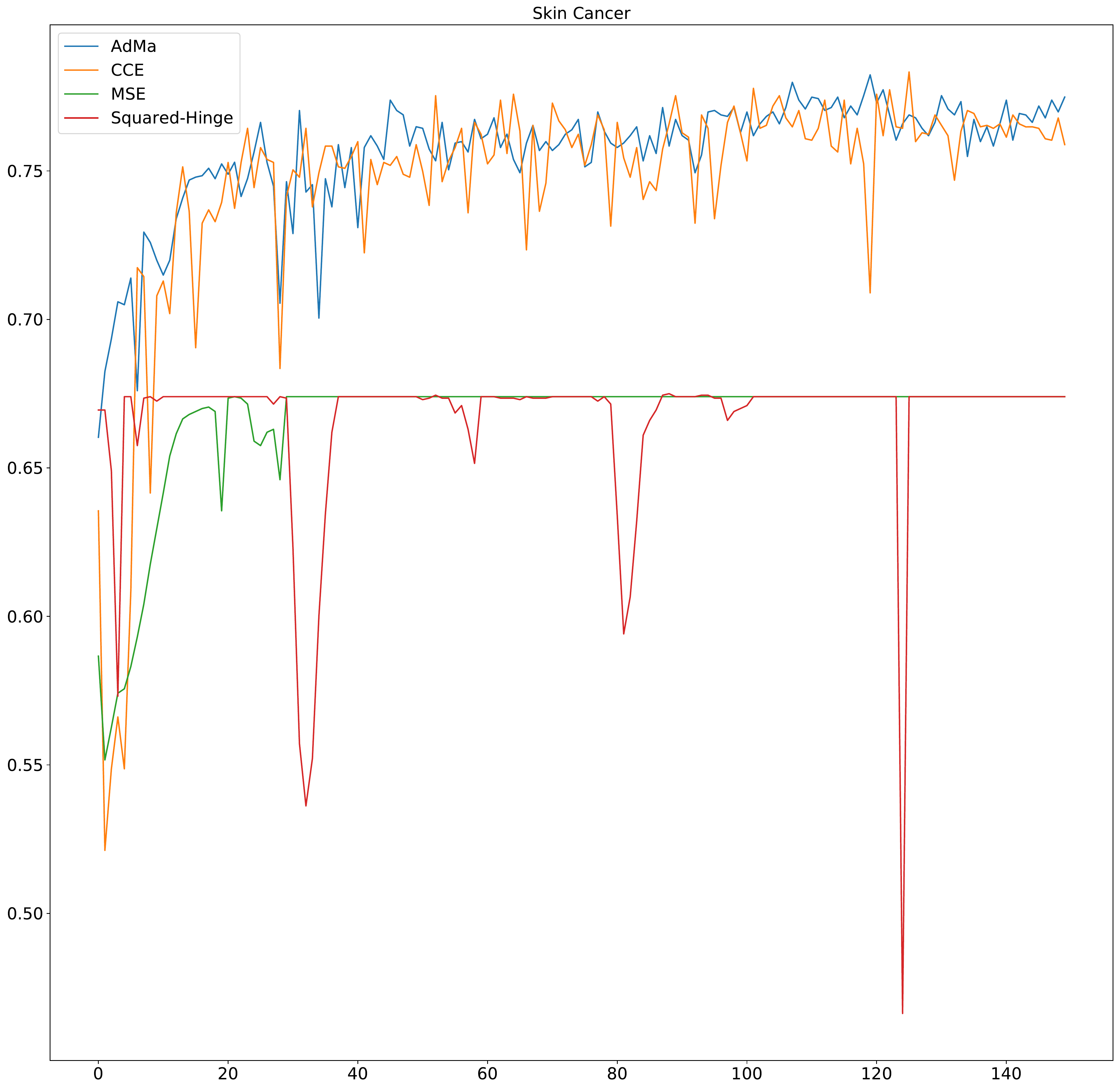}\par 
    \includegraphics[width=\linewidth]{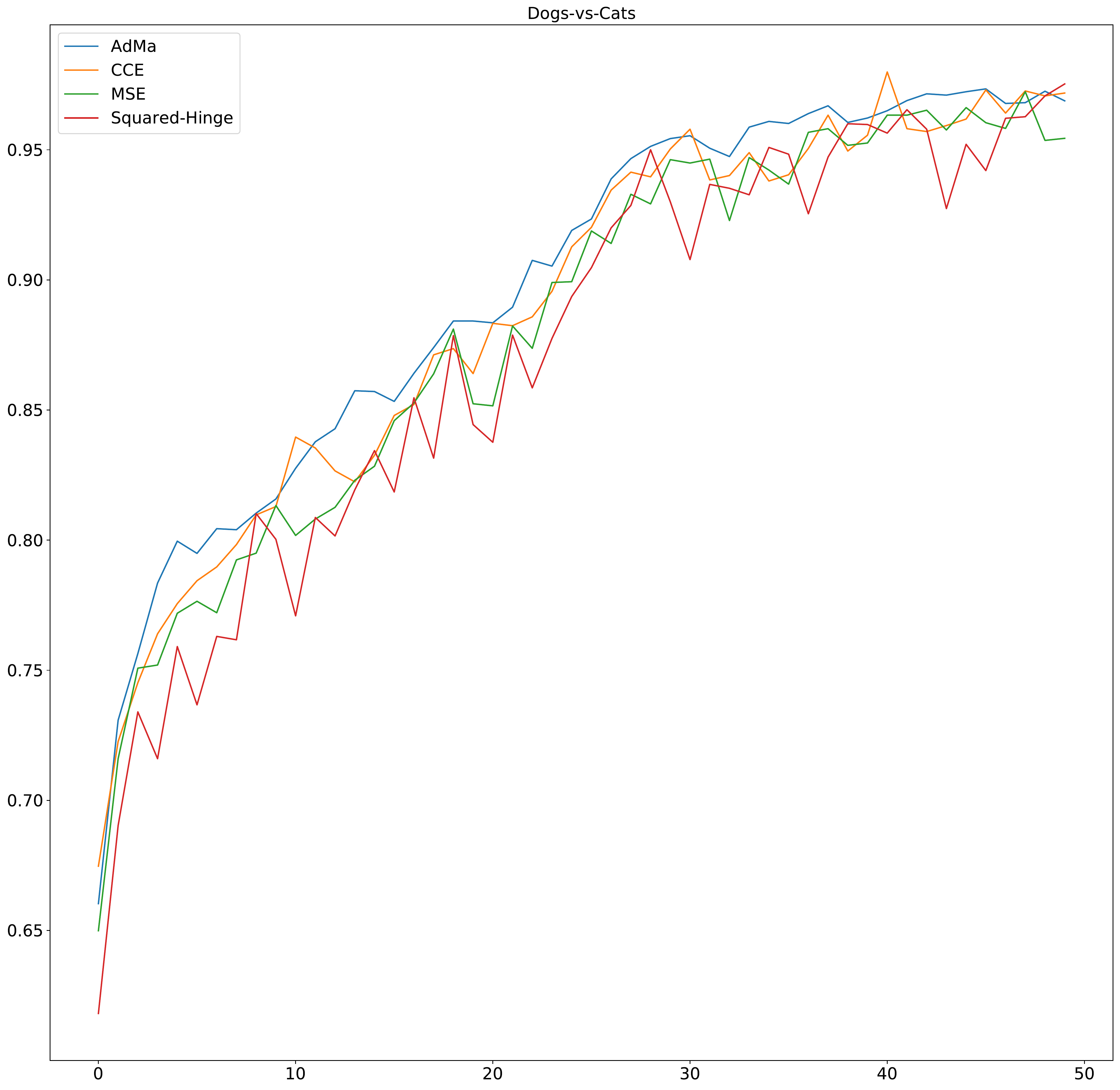}\par 
\end{multicols}
\caption{The above set of figures depict a typical tracing of learning in the model. The order from left-to-right row-wise is as follows: MNIST, Fasion MNIST, CIFAR10, CIFAR100, SVHN, Flower Recognition, Skin Cancer, Dogs-vs-Cats}
\label{fig:3}
\end{figure*}

Since, flexibility is the central theme of the function, we analysed and tested the same for our function on plethora of settings. In this section, we mention about all these  settings and discuss the characteristics that show the most significant role-play in the performance with our loss function. Initially, the ConvNet is trained one by one on each dataset using Adam optimizer as it is adaptable and has a faster convergence rate than SGD. These experiments had shown that Adma was able to nearly parallel the performance of CCE and even surpass it by a marginal value in some cases (refer tab. \ref{tab2}, \ref{tab1} \& \ref{fig:3}). One thing that can be primarily noticed from both the figures and tables is that the prime contention during all of the experimentation was observed to be only between Adma and the CCE. For this reason, we did each of our experiments under the settings when the performance with CCE was found to be optimal and then try to surpass that accuracy by adjusting the \textit{scaling factor} of Adma. Otherwise, there are certain conditions under which Adma's accuracy surpasses all of the other cost function's accuracies by a marginal value. We shall be mentioning about these conditions and the reason behind this behaviour as we proceed through the discussion. Thus, the primary objective of this section is to discuss about the observations and results obtained during the experiments and further in developing an intuition towards using the function to improve the results. 

\begin{table}[H]
\begin{tabular}{|c|c|c|c|c|}
\hline
                   & Adma                                         & CCE                               & Squared-Hinge & MSE                               \\ \hline
MNIST              & \textbf{(a=0.2624) 99.34\%} & 99.30\%                           & 98.45\%       & 98.74\%                           \\ \hline
FASHION-MNIST      & \textbf{(a=0.2580) 91.48\%} & 91.42\%                           & 86.13\%       & 88.76\%                           \\ \hline
CIFAR10            & (a=0.2622) 86.76\%                           & \textbf{87.28\%} & 82.41\%       & 85.42\%                           \\ \hline
CIFAR100           & (a=0.2820) 57.60\%                           & \textbf{57.76\%} & 20.23\%       & 35.19\%                           \\ \hline
Flower Recognition & (a=0.2625) 77.70\%                           & \textbf{77.78\%} & 55.93\%       & 58.34\%                           \\ \hline
Skin Cancer        & \textbf{(a=0.2625) 77.08\%} & 75.39\%                           & 66.03\%       & 67.40\%                           \\ \hline
SVHN               & (a=0.2612) 95.14\%                           & \textbf{95.67\%} & 92.21\%       & 93.12\%                           \\ \hline
Dogs vs. Cats      & (a=0.2451) 98.31\%                           & 97.67\%                           & 98.53\%       & \textbf{98.64\%} \\ \hline
\end{tabular}
\caption{Accuracies achieved with regularization and Adam as an optimizer.}
\label{tab1}
\end{table}

\begin{table}[H]
\begin{tabular}{|c|c|c|c|c|}
\hline
                   & Adma                        & CCE              & Squared-Hinge & MSE     \\ \hline
MNIST              & (a=0.2712) 98.54\%          & \textbf{98.78\%} & 96.87\%        & 97.46\%  \\ \hline
FASHION-MNIST      & (a=0.2729) 83.21\%          & \textbf{84.31\%} & 79.71\%       & 78.06\% \\ \hline
CIFAR10            & \textbf{(a=0.2809) 83.74\%} & 83.45\%          & 78.46\%       & 82.03\% \\ \hline
CIFAR100           & (a=0.3110) 31.31\%          & \textbf{40.91\%} & 12.39\%       & 20.23\% \\ \hline
Flower Recognition & (a=0.2762) 54.68\%          & \textbf{57.32\%} & 45.15\%       & 49.41\% \\ \hline
Skin Cancer        & \textbf{(a=0.2918) 71.11\%} & 68.05\%          & 37.58\%       & 63.96\% \\ \hline
SVHN               & (a=0.3007) 69.92\%          & \textbf{74.12\%} & 49.01\%       & 53.46\% \\ \hline
Dogs vs. Cats      & \textbf{(a=0.2727) 88.16\%} & 85.29\%          & 79.98\%       & 78.45\% \\ \hline
\end{tabular}
\caption{Accuracies achieved with regularization and SGD as an optimizer.}
\label{tab2}
\end{table}


\begin{table}[H]
\begin{tabular}{|c|c|c|c|c|}
\hline
                   & Adma                                         & CCE                               & Squared-Hinge                     & MSE                               \\ \hline
MNIST              & \textbf{(a=0.2624) 99.51\%} & 99.42\%                           & 99.29\%                           & 99.46\%                           \\ \hline
FASHION-MNIST      & \textbf{(a=0.2481) 92.14\%} & 91.86\%                           & 90.15\%                           & 91.56\%                           \\ \hline
CIFAR10            & \textbf{(a=0.2489) 85.08\%} & 84.92\%                           & 82.92\%                           & 83.81\%                           \\ \hline
CIFAR100           & (a=0.3080) 57.74\%                           & \textbf{58.65\%} & 34.22\%                           & 43.28\%                           \\ \hline
Flower Recognition & \textbf{(a=0.2671) 80.72\%} & 80.57\%                           & 53.36\%                           & 58.87\%                           \\ \hline
Skin Cancer        & \textbf{(a=0.2625) 78.12\%} & 75.77\%                           & 62.12\%                           & 66.95\%                           \\ \hline
SVHN               & (a=0.2631) 95.80\%                           & 95.56\%                           & 95.45\%                           & \textbf{95.88\%} \\ \hline
Dogs vs. Cats      & (a=0.2653) 96.88\%                           & 97.18\%                           & \textbf{97.53\%} & 95.44\%                           \\ \hline
\end{tabular}
\caption{Accuracies achieved without regularization and Adam as an optimizer.}
\label{tab3}
\end{table}

\begin{table}[H]
\begin{tabular}{|c|c|c|c|c|}
\hline
                   & Adma                                         & CCE                               & Squared-Hinge & MSE     \\ \hline
MNIST              & (a=0.2628) 98.97\%                           & \textbf{99.07\%} & 97.06\%       & 98.37\% \\ \hline
FASHION-MNIST      & \textbf{(a=0.2654) 87.75\%} & 86.16\%                           & 81.45\%       & 85.78\% \\ \hline
CIFAR10            & (a=0.2807) 83.99\%                           & \textbf{84.12\%} & 80.03\%       & 82.71\% \\ \hline
CIFAR100           & (a=0.3071) 25.31\%                           & \textbf{45.13\%} & 19.57\%       & 32.14\% \\ \hline
Flower Recognition & (a=0.2697) 59.92\%                           & \textbf{65.25\%} & 50.87\%       & 56.15\% \\ \hline
Skin Cancer        & \textbf{(a=0.2789) 76.36\%} & 72.46\%                           & 47.81\%       & 65.42\% \\ \hline
SVHN               & \textbf{(a=0.2986) 68.49\%} & 66.25\%                           & 59.58\%       & 56.78\% \\ \hline
Dogs vs. Cats      & \textbf{(a=0.3118) 93.24\%} & 92.71\%                           & 83.94\%       & 86.64\% \\ \hline
\end{tabular}
\caption{Accuracies achieved without regularization and SGD as an optimizer.}
\label{tab4}
\end{table}

Firstly, with L2 regularization technique being applied on the model, the performance with Adma is highest on half the datasets namely MNIST, FASHION-MNIST, Flower Recognition and Skin Cancer. The highest margin is observed for the Skin Cancer dataset. For the rest of the datasets, difference is minimal. Therefore, it may not be incorrect to say that for some alternative conditions CCE may be better than Adma. However, One distinguishing characteristic of the Skin Cancer dataset as compared to other datasets is that it has a rectangular shape. This hints that Adma might be better with the irregular shaped data. Further, shifting from with regularization to without regularization, Adma takes the lead two more datasets. This shows that when the model is used unequivocally, without any implicit nourishment, Adma enables the learning more appropriately. On the other hand, CCE even though it starts out with lesser accuracy, later starts learning more rapidly.

The ConvNet, with Adma as a loss function, exhibits a state-of-the-art performance on not only the common but also on somewhat peculiar datasets. For eg. CIFAR100 is a 100 label dataset with only 500 images per category for training. On that, the model outputs the accuracy of approx. 57\%, which remains invariant even when the regularization is removed. The performance in this case is very comparable to the CCE and is remarkably higher than the other two cases. Next, on the flower dataset which is small dataset with high resolution RGB images, the performance with Adma remains the same as on the other datasets i.e. highly contentious to CCE. In fact, without the support of regularization, Adma performs the best amongst all. And this accuracy is the highest registered on this dataset taking all of the explored hyperparameter search space into consideration. Lastly, on the SVHN dataset also, where the task is to correctly classify the centre digit from the images containing 5 digits in total, Adma gives the reasonable results. Thus, from the performance on all these vivid datasets, it can be derived that Adma is a reliable and a genuine loss function. 

Thereafter, we also believe that all of the graphs presented in Fig. \ref{fig:3} indicate some common things about Adma. The first behaviour that can be roughly observed is that Adma almost always starts out better than the rest of the functions. And that also by a significant margin. For eg. on the most common, MNIST dataset, Adma was the only function that had achieved an accuracy of approx. 98\% during its first epoch, when CCE ~96.5\%, MSE ~95.5\% and Squared Hinge was ~93.5\%. The another thing the figures are evident of is that throughout training, Adma learns gradually and is relatively more stagnant (showing less variance) than other cost functions. From this it can be interpreted that Adma, more efficiently quantifies to the noise in the data and is in true sense the adaptable loss function. Thus, the above two cruxes point towards the fact that model shows higher convergence rate with Adma as a loss function. One another thing to note here is that even though the performance of CCE is emulating (or higher in some cases) to Adma, comparatively higher variance is observed for the CCE. Further, talking about the other loss functions, there are some datasets namely Fashion MNIST, CIFAR100, Flower Dataset and Skin Cancer that advocate Squared Hinge as worst performing function amongst all. In fact, the learning seems to be completely impaired for CIFAR100 and Skin Cancer datasets. Next, the MSE has all the same but quite better performance than Squared Hinge in terms of accuracy and variance. Still, with MSE as a loss function, model struggles to achieve the accuracies similar to that of Adma or CCE on all the datasets except for SVHN and Dogs-vs-Cats. Thus, in overall sense, collectively considering accuracy, variance and convergence rate, Adma clearly seems to be the most suitable choice. 


With SGD as an optimizer, it was quite a hard time finding the optimal scaling factor. And therefore a competent accuracy. It was observed from the experiments that Adma when used with SGD requires higher scaling factor to rise to satisfactory accuracy. Here, the model was compiled with SGD by keeping the default value of the learning rate i.e. $1\times10^{-3}$. Therefore, to lift off the performance of Adma, we tried incorporating the more complex functionalities such as enabling Nestorov Momentum, learning rate annealer etc. However, the similar proportion of upliftment was observed for almost all of the loss functions. But, amid such readjustments, when we reduced the learning rate to as low as $2\times10^{-4}$, exceeded performance was recorded with Adma than the rest of the functions. On applying the same technique with Adam, an identical behavior was observed and Adma achieved leading accuracy on SVHN, CIFAR10 and Flower Recognition datasets as well. Thus, for some reduced learning rate, ConvNet with Adma as loss function was successfully able to overshadow the performance recorded with all the other loss functions. One thing that can be derived for this is that for non-adaptive optimizers like SGD, getting higher accuracy with our loss function seems to be an interplay of the chosen learning rate and scaling factor. Ultimately for some combination of these two we were able to achieve the accuracy higher or parallel to that with other loss functions. However, it is important to mention here that the accuracies observed under this condition of lowered learning rate, are not the benchmarking accuracies. Also, the lowered learning rate on SGD may lead to prolonged convergance. Also note that in this particular experimentation, the objective was to manifest the conditions that demonstrate the possibility of Adma to fully exceed the prevalent functions' standards so that they may prove as a perceptive guidelines for the improvements in future. 

We would like to present another such setting--test-train split that hints Adma as an outlier as compared to other functions. The test-train split of the standard datasets is fixed and thus we do not tamper the same. In this particular scenario, we regard the non-standard datasets as the datasets that are not made available by the deep learning frameworks itself. However, for the non standard datasets, the proportion of the test-train split can be easily changed. By default, for each of the non standard datasets, the test-train split was 0.8. With this setting, Adma already emulates the performance of all objective functions on flower recognition and skin cancer datasets. Further, on reducing the test-train split to 0.75, all of the cost functions' performances was dropped. But, the proportion by which the performance of Adma function had dropped was the least. Finally on bringing down the test-train split to as low as 0.6, Adma was able to lead the CCE function on an average by approzimately 4\%.  
This means that when you have lower amount of data to train with and subsequently, higher amount of data to validate on, the Adma seems to be a significantly better fit.
Moreover, under this set of circumstances the training accuracy with CCE was always higher than rest of the datasets but on the validation data the accuracy with Adma was found to be higher than all other functions. One interpretation for this can be that Adma enables a more generalized learning of the data and is less prone to overfitting. Thus, this is another result confirming the robustness of the Adma besides the one achieved upon the removal of regularization.

Now we briefly mention about the rest of the environments and settings that were considered for testing our loss function and still did not turn out to be so lucrative. One substitution was of various activation functions. The default activation function in all of the experiments done is ELU. Therefore, we replace the same with the activation functions ReLU, it's variant Leaky ReLU and Maxout. Nevertheless, no significant exception was observed for Adma. Similarly, we also tried tampering the internal structure of the ConvNet such as adjusting the dropout rate, changing window size and stride length etc. but no significant improvement over other functions was observed for Adma. Lastly, to validate the performance on state-of-the-art pretrained model, we shift to ResNet34. However, due to lack of time, we only tested the model on 4 standard datasets--MNIST, Fashion MNIST, CIFAR10, CIFAR100. Also, due to the limitation upon the computational resources available, the model is trained directly by inheriting the pre-trained weights and not from scratch. Here, we only allow to train the top 5 layers of the model. Thereafter, all of the same behavior and the results were observed on ResNet34 model as well. To say the least, the model successfully achieves the state-of-the-art performance on all the tested datasets. MNIST has an error rate of 0.025\%, Fashion MNIST of around 6.5\% CIFAR10 $\sim$9\% and CIFAR100 $\sim$38\%. Moreover, we also observed highest first epoch accuracy depicting loftier finessing power of the function. However, on this model rejigging the learning rate did not eventuate any substantial boost. We suppose that this is because only the top 5 layers were allowed to be trained. Hence the values of the errors to be propogated or update steps were plenteously large for all 5 layers to be trained.




\section{Conclusion} \label{six}

With this, we shall conclude with proposing our novel, flexible loss function and put forth the concluding remarks and future scope of this work. 
It is shown by empirical demonstration that the proposed loss function, motivated by the idea of flexibility, can very emulate or even surpass the performance of prevalent loss functions. Here, we have presented a rather more detailed analysis on the experimental validation section than the theoretical analysis because we feel that it shall be more helpful guide to the community for further analysis and improvements. And also several works in deep learning mention the theoretical results to be largely deviating from experimental results. 

To the best of our knowledge, this is the pioneering work on the development of purely flexible loss function for neural networks. The importance of such function is fostered by the biologically plausibility of the variable loss elicitation in our prefrontal cortex. The presented work broaches the fundamental reasons for it to be noteworthy enough to 
reinstigate an active research on the improvement of loss functions. 
We also feel that somehow an extensive research on the incorporation of the flexibility, adaptability and dynamicity criteria in the structure of neural networks can cause a paradigm shift and help us quickly achieve an unparalleled performance in Artificial General Intelligence. 

One may notice that this work questioned one of the most commonly held belief why CCE is widely accepted-- the logarithm undoes the exponentialized softmax output units so that the gradients, while back-propagation show linear nature. Whereas, in Adma loss function the exponentialized output units are even further exponentialized, and still similar performance is achieved. This suggests that one should not take even the Adma function as it is. Rather, one should question its structure as well. For eg. trying out different base values other than $\exp$ since, it only has a numerical importance. Due to such precocious nature of the proposed work, we intentionally miss out on providing any solid theoretical framework for the function. A theoretical framework may limit the structure of proposed function whereas the function requires to be explored more which we believe can be done best by experimentative exploration rather than theoretical proofs. 

In general, the intuition about the optimal value of scaling factor, just like many other values of hyper-parameters can be developed as one proceeds with the experimentation. However, the future work can be to turn the same into a parameter and learn the same during the training process. Further, the function can even be investigated with more complicated networks such as generative models, image segmentation networks etc. where the design and choice of loss function plays a more significant role.


Thus, all of the above facts insinuate that the proposed Adma loss function is already able to deliver state-of-the-art performance, alongside leaving a room for significant exploration and further possibilities. This may henceforth introduce a new paradigm of flexibility and adaptability in deep learning research.

\bibliographystyle{IEEEtran}
\bibliography{as}
%




%







\end{document}